\documentclass[runningheads]{llncs}

\usepackage{eccv}

\usepackage{eccvabbrv}

\usepackage{graphicx}
\usepackage{booktabs}

\usepackage[accsupp]{axessibility}  

\usepackage[pagebackref,breaklinks,colorlinks,citecolor=eccvblue]{hyperref}

\usepackage{wrapfig}

\usepackage{hyperref}

\usepackage{orcidlink}

\begin{document}

\title{WearWow: Native 2K Multi-Garment Virtual Try-On via Adaptive Token Packing and Preference Alignment} 

\titlerunning{WearWow}

\author{Xujie Zhang\inst{1}\setcounter{footnote}{0}\thanks{Equal contribution.} \and
Runyan Du\inst{2}$^{\star}$ \and
Song Chang\inst{2} \and
Jiang Li\inst{2} \and
Dongliang Shao\inst{2} \and
Liping Wu\inst{2} \and
Wei Luo\inst{2} \and
Xiaochao Qu\inst{2} \and
Luoqi Liu\inst{2} \and
Xiaodan Liang\inst{1}\setcounter{footnote}{3}\thanks{Corresponding author.}}

\authorrunning{X. Zhang. et al.}

\institute{Shenzhen Campus of Sun Yat-sen University, China \\
\email{zhangxj59@mail2.sysu.edu.cn, liangxd9@mail.sysu.edu.cn} \and
Meitu Lab, China \\
\email{\{dry, cs2, lj28, sdl, wlp1, lw20, qxc, llq5\}@meitu.com}}

\maketitle
{ 
    \centering
    \captionsetup{type=figure}
    \includegraphics[scale=0.65]{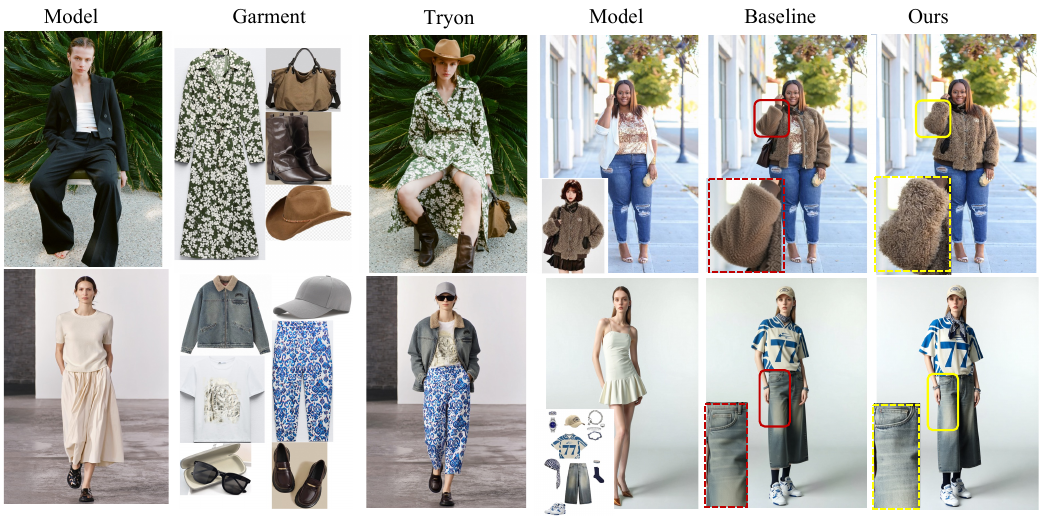}

    \captionof{figure}{\textbf{Native 2K Multi-Garment Try-On with WearWow.} (Left) Mask-Free Joint Synthesis: WearWow seamlessly composes diverse categories (garments, footwear, accessories) with strict topological alignment. (Right) Tactile Realism: While standard baselines suffer from severe over-smoothing, our framework explicitly restores micro-level high-frequency textures (e.g., curly fleece, denim), overcoming plastic-like degradation to achieve unprecedented material fidelity.}
    \label{fig:teaser}
}

\begin{abstract}
Synthesizing native 2K multi-garment virtual try-on is a formidable frontier in digital fashion, critically bottlenecked by two fundamental limitations: the $\mathcal{O}(N^2)$  memory explosion induced by 2k conditions, and the spectral bias of diffusion models that over-smooths high-frequency fabric details. We present WearWow, an end-to-end, mask-free generative framework that pioneers ultra-high-resolution multi-garment synthesis. To mitigate the memory explosion , we propose Adaptive 2D Token Packing (ATP). ATP leverages inherent garment sparsity to algorithmically pack heterogeneous items onto a unified 2D canvas and prune uninformative background tokens, minimizing the effective sequence length and subsequent memory overhead while rigorously preserving 2D spatial priors. To rectify texture degradation, we introduce the Multi-dimensional Try-on Reward (MTR) system. MTR synergizes a Semantic Guidance Reward to explicitly drive tactile restoration with a Cloth Distribution Reward to implicitly anchor the physical distribution, a joint formulation that effectively mitigates the severe reward hacking. Furthermore, we curate WearWow-2K, an extreme-quality dataset comprising native 2K triplets, providing physically correct spatial interactions that naturally empower the model's mask-free generation. Extensive experiments demonstrate that WearWow establishes a new state-of-the-art, exceeding existing commercial baselines in native 2K multi-garment synthesis.

\keywords{Multi-garment Try-on \and Native 2K Resolution \and Mask-free Generation \and Preference Optimization}
\end{abstract}

\section{Introduction}
\label{sec:intro}

Image-based Virtual Try-On (VTON)~\cite{ge2021parser, bai2022single, xie2023gp, morelli2023ladi, gou2023taming} is undergoing a profound paradigm shift: evolving from the macroscopic pursuit of ``structural alignment''---typically constrained to 512p or 1K resolutions to ensure correct pose warping---towards the industrial-grade demand for ``tactile realism'' at ultra-high resolutions (2K and beyond). Driven by the rapid evolution of generative diffusion models, recent state-of-the-art architectures have demonstrated remarkable success in preserving complex human poses and single-garment semantics. However, when transitioning to authentic commercial deployments, which strictly necessitate native 2K resolution combined with multi-garment joint generation (e.g., simultaneously synthesizing inner-wear, outerwear, and bottoms) and mask-free inference, existing end-to-end frameworks encounter severe systemic bottlenecks. As the generative objective fundamentally shifts from merely ``wearing it correctly'' to ``rendering authentic physical textures,'' current paradigms face a significant degradation, bottlenecked by both physical hardware limitations and theoretical generative flaws.

Specifically, scaling diffusion-based VTON to the 2K multi-garment regime is obstructed by two primary scaling barriers. The first is the computational bottleneck driven by sequence length. In modern U-Net or Diffusion Transformer backbones, the computational complexity of self-attention mechanisms scales quadratically with the total number of visual tokens. Alongside the fixed base tokens $N_{\text{base}}$ required for the target and model images, naively concatenating $K$ high-definition reference garments (each containing $N$ tokens) directly expands the total sequence length to $N_{\text{base}} + K \cdot N$. This triggers an intractable computational cost of $\mathcal{O}((N_{\text{base}} + K \cdot N)^2)$ and a catastrophic spatial memory explosion, rendering native 2K multi-garment training fundamentally infeasible. Consequently, existing methods are forced to rely on suboptimal cascaded super-resolution pipelines, which inevitably introduce structural hallucinations and disrupt native spatial coherence.

The second barrier is the generative quality degradation caused by high-frequency attenuation. Even if memory constraints are artificially circumvented, generative models face a fundamental signal processing flaw at extreme spatial scales. The conventional $L_2$-norm (MSE) denoising objective exhibits an inherent spectral bias and a mean-seeking property. While relatively imperceptible at 1K resolution, this bias acts as an aggressive low-pass filter at the 2K spatial dimension. It systematically smooths out the microscopic high-frequency spatial components crucial for rendering complex textiles. As a result, materials heavily dependent on high-frequency details, such as coarse wool, heavy knits, and rugged denim, suffer from severe material degradation, deteriorating into overly smoothed, plastic-like synthetic surfaces that fail to capture tactile authenticity.

To systematically address these computational and theoretical bottlenecks, we propose WearWow, a unified framework specifically engineered for mask-free, native 2K multi-garment virtual try-on. The core of our framework is driven by a sequential design philosophy: first, making extreme-resolution training computationally feasible; and second, ensuring the generated textures achieve industrial-grade tactile realism. To achieve the first objective and fundamentally mitigate the computational explosion associated with multi-condition processing , we introduce Adaptive 2D Token Packing (ATP). By exploiting the inherent spatial sparsity of flat-lay garment images, ATP dynamically prunes uninformative background regions and packs heterogeneous visual tokens into a highly compact 2D latent matrix. Crucially, ATP preserves the original 2D spatial topology through rigorously mapped 2D positional encodings. By tightly packing valid regions and aggressively pruning background voids, this spatial reorganization drastically slashes the effective sequence length and subsequent memory footprint, enabling end-to-end joint training on standard hardware while mathematically guaranteeing spatial inductive biases. 

To achieve the second objective and counteract the spectral attenuation inherent in standard diffusion objectives, we formulate the Multi-dimensional Try-on Reward (MTR) pipeline. MTR systematically shifts the optimization paradigm from purely latent-space regression to fine-grained preference alignment. Specifically, MTR introduces a Cloth Distribution Reward (CDR) to implicitly anchor the generative trajectory to high-fidelity physical data distributions, and a Semantic Guidance Reward (SGR) to explicitly drive tactile restoration via contrastive textual anchors. This dual-dimensional alignment effectively neutralizes the spectral bias of MSE, physically restoring the micro-level structural integrity of complex fabrics without succumbing to severe reward hacking.

The advancement of ultra-high-resolution digital fashion is further hindered by the severe scarcity of high-fidelity, multi-garment paired data at extreme spatial scales. To establish a rigorous benchmark and enable true mask-free generative capabilities, we curate and release WearWow-2K. This large-scale dataset features native 2K paired images with synthesized agnostic models, providing flawless high-frequency texture gradients and physically correct spatial interactions that naturally empower the model's target person mask-free generation. In summary, our core contributions are four-fold:
\begin{itemize}
    \item We propose \textbf{WearWow}, the effective end-to-end multi-garment virtual try-on framework operating natively at 2K resolution, effectively advancing the task from macroscopic structural warping to micro-level tactile rendering.
    
    \item We introduce \textbf{Adaptive 2D Token Packing} , a spatial representation mechanism that mitigates the computational bottleneck by minimizing the effective sequence length while drastically slashing the sequence footprint and strictly preserving 2D spatial priors.
    
    \item We formulate the \textbf{Multi-dimensional Try-on Reward pipeline}. It effectively counteracts the spectral bias of diffusion models to physically rescue high-frequency fabric details and overcome plastic-like degradation.
    
    \item We release the \textbf{WearWow-2K dataset } featuring synthesized agnostic multi-garment models to unlock mask-free generation, and establish 2K resolution evaluation dimensions for ultra-HD digital fashion research.
\end{itemize}

\section{Related Work}

\textbf{Single-Garment Virtual Try-on.}
Early virtual try-on paradigms predominantly relied on Generative Adversarial Networks (GANs)~\cite{dong2019towards,yang2020towards,dong2022dressing,lee2022high,xie2022pasta,he2022fs_vton,xie2023gp,xie2021vton,wang2018cpvton,Minar2020CP-VTON+}, typically employing a two-stage pipeline consisting of an explicit warping module followed by a blending generator. While effective for macroscopic pose alignment, GAN-based architectures inherently struggle with synthesizing complex, high-frequency fabric textures and scaling to ultra-high resolutions.

Recently, Diffusion Models~\cite{rombach2022high} have revolutionized the field by treating try-on as a conditional generation or inpainting task~\cite{ramesh2022hierarchical,yang2023paint,huang2023composer,chen2023anydoor}. Methods such as TryonDiffusion~\cite{zhu2023tryondiffusion} introduced implicit warping mechanisms to better handle fabric deformations. Subsequent state-of-the-art frameworks---including OOTDiffusion~\cite{xu2024ootdiffusion}, IDM-VTON~\cite{choi2024idmvton}, Any2AnyTryOn~\cite{guo2025any2anytryon}, IMAGDressing~\cite{shen2024IMAGDressing-v1}, Leffa~\cite{zhou2024leffa}, PromptDresser~\cite{hertz2022prompt}, cascade~\cite{li2025cascaded}and CatVTON~\cite{chong2024catvtonconcatenationneedvirtual}---have progressively refined appearance preservation through advanced cross-attention strategies and tailored condition encoders. Despite their remarkable success in preserving complex human poses and single-garment semantics, these architectures are intrinsically designed for single-item scenarios. When naively scaled to native 2K resolutions, they inevitably encounter the spectral bias of the conventional MSE objective, leading to over-smoothed, ``plastic-like'' textures that fail to satisfy industrial-grade tactile realism.

\textbf{Multi-Garment Joint Synthesis.}
Compared to single-item VTON, multi-garment synthesis presents a significantly more daunting challenge due to the complex spatial occlusions and the exponential surge in condition tokens. While some early works explored compositional try-on via cyclical or sequential generation~\cite{cui2021dressing,xie2022pasta, Lin2025DreamFit, zhang2024mmtryonmultimodalmultireferencecontrol}, they inherently suffer from severe error accumulation and unnatural layer blending. More recently, OminiTry~\cite{feng2025omnitry} pioneered the exploration of multi-garment inputs within a unified diffusion framework. However, due to the quadratic ($\mathcal{O}(N^2)$) memory explosion of self-attention mechanisms, there remains a severe lack of dedicated, open-source vertical baselines capable of executing native 2K multi-garment joint inference without catastrophic memory failures (OOM).

Consequently, the community has often looked toward state-of-the-art Large Vision-Language Models (LVLMs) and commercial generative APIs (e.g., QWen-image~\cite{wu2025qwen}, Seedream4.0~\cite{seedream2025seedream}, GPT-image 1.5~\cite{achiam2023gpt}, KeLingv3~\cite{team2025kling}) to handle extreme combinatorial complexity. While these foundation models have demonstrated astonishing semantic understanding and zero-shot compositional capabilities, they remain generalized text-to-image engines. They frequently suffer from cross-garment feature leakage, struggle to maintain spatial coherence under complex occlusion, and fail to render authentic material fidelity. 

In stark contrast, our proposed WearWow is specifically engineered for this extreme frontier. By fundamentally resolving the multi-condition memory bottleneck via Adaptive 2D Token Packing (ATP) and explicitly rescuing high-frequency tactile textures through the Multi-dimensional Try-on Reward , WearWow establishes a robust, highly efficient vertical solution for ultra-high-resolution digital fashion.

\section{Methodology}
\label{sec:methodology}

\begin{figure}[t]
  \centering
  \includegraphics[width=1.0\textwidth]{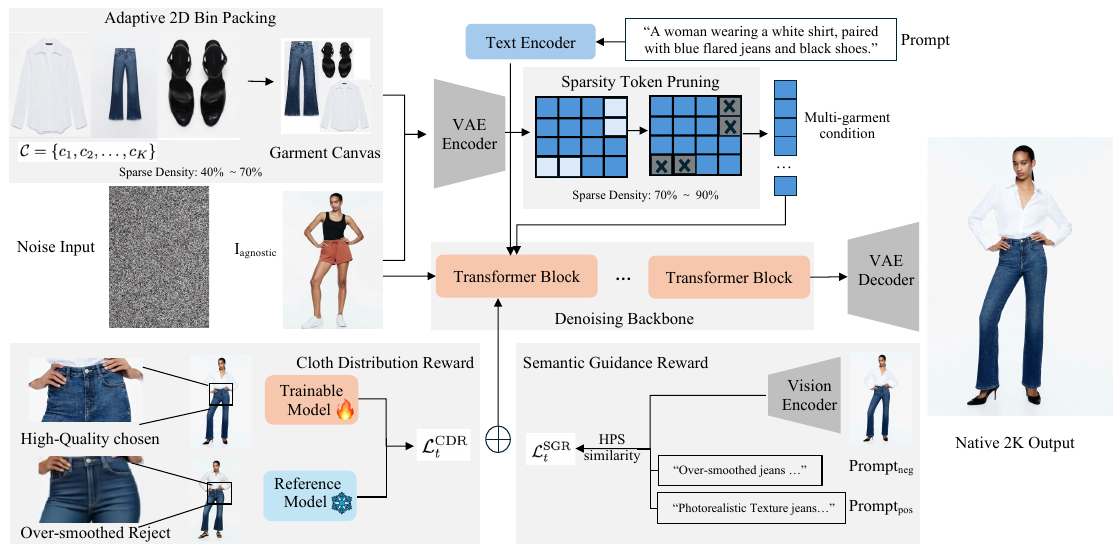}
  
  \caption{The overall framework of WearWow. Our pipeline enables mask-free, native 2K multi-garment virtual try-on through two core mechanisms. \textbf{(Top) Adaptive 2D Token Packing:} Given a set of reference garments $\mathcal{C}$, ATP algorithmically packs them onto a unified spatial canvas. Following VAE encoding, a Sparisity Token Pruning mechanism aggressively discards background void tokens. This drastically minimizes the effective sequence length to mitigate the memory explosion, while strictly preserving 2D positional encodings. \textbf{(Bottom) Multi-dimensional Try-on Reward:} To overcome the spectral bias of standard SFT, MTR fine-tunes the generative trajectory. It synergizes a Cloth Distribution Reward (CDR) that penalizes over-smoothed artifacts to anchor physical distributions, and a Semantic Guidance Reward (SGR) that leverages HPS-driven contrastive textual anchors ($Prompt_{pos}, Prompt_{neg}$) to explicitly restore high-frequency tactile realism.}

  \label{fig:framework}

\end{figure}

\subsection{Overview and Problem Formulation}
\label{subsec:overview}

Native 2K multi-garment virtual try-on demands a unified generative paradigm capable of disentangling complex spatial interactions while strictly bounding computational overhead. Formally, WearWow accepts a triplet of conditional inputs: (1) a clothing-agnostic target person image $I_{\text{agnostic}}$; (2) a heterogeneous set of $K$ high-resolution reference garments $\mathcal{C} = \{c_1, c_2, \dots, c_K\}$; and (3) a layer-aware textual prompt $\mathcal{P}$. To enable mask-free generation, we construct the WearWow-2K data engine using a ``Real Ground-Truth, Synthesized Condition'' paradigm. We utilize authentic 2K multi-garment photographs as the absolute target $\hat{I}_{\text{target}}$ to preserve pristine high-frequency textures, while inversely synthesizing $I_{\text{agnostic}}$ via a robust try-off pipeline. This formulation forces the model to implicitly learn physical depth ordering without relying on precise manual masking prior to inference (dataset details are deferred to the supplementary).

As illustrated in Figure~\ref{fig:framework}, successfully scaling this formulation into the native 2K regime requires addressing two sequential hurdles: computational feasibility and generative quality. Our framework tackles these through two core components. First, Adaptive 2D Token Packing (ATP, Section~\ref{subsec:atp}) effectively mitigates the spatial memory bottleneck, providing a practical solution for how to train at extreme resolutions. Second, building upon this computationally feasible foundation, the Multi-dimensional Try-on Reward (MTR, Section~\ref{subsec:mtr}) counteracts the inherent spectral bias of diffusion models, addressing how to train effectively to restore authentic tactile realism.

\subsection{Adaptive 2D Token Packing (ATP)}
\label{subsec:atp}
In authentic virtual try-on scenarios, high-resolution reference garments inherently contain a massive proportion of uninformative background pixels. In a standard Transformer-based generative backbone, the input sequence consists of the fixed-length base tokens $N_{\text{base}}$ (derived from the noisy target and the clothing-agnostic model) and the conditional garment tokens. Naively injecting $K$ reference garments, each producing $N$ visual tokens, extends the total sequence length to $N_{\text{base}} + K \cdot N$. Given the inherent quadratic complexity of self-attention mechanisms, this naive concatenation triggers a prohibitive computational cost of $\mathcal{O}((N_{\text{base}} + K \cdot N)^2)$. As $K$ increases, this geometric explosion leads to catastrophic VRAM consumption and severe cross-garment spatial interference. To mitigate this without compromising high-frequency structural integrity, we propose the Adaptive 2D Token Packing (ATP) mechanism.

\textbf{Adaptive 2D Bin Packing.}
Crucially, ATP does not alter the inherent quadratic nature of self-attention but strategically minimizes the conditional sequence length. Rather than concatenating conditions in 1D---which arbitrarily destroys spatial priors---ATP formulates reference injection as an algorithmic 2D bin packing optimization. The $K$ heterogeneous items are dynamically scaled based on geometric proportions and tightly packed onto a unified 2D spatial canvas with a fixed token capacity $L$ (where $L \leq K \cdot N$). This spatial recombination instantly decouples the overall complexity from the quadratic scaling of $K$, strictly bounding the theoretical cost to $\mathcal{O}((N_{\text{base}} + L)^2)$. Empirically, this packing dramatically increases effective foreground token utilization from a baseline of approximately 40\% (inherent in standard padded batching) to nearly 70\%, while strictly isolating spatial coordinates to prevent feature leakage.

\textbf{Sparsity Token Pruning and Fallback.}
To push inference efficiency to the physical limit, ATP further exploits the inherent spatial sparsity of the unified canvas. Despite optimal 2D bounding-box packing, the canvas inevitably retains uninformative blank padding due to the irregular topology of garments. Utilizing precise binary valid-region masks, ATP explicitly prunes and discards $m$ visual tokens corresponding to these background void spaces post-patchification. Consequently, the actual conditional sequence length processed by the transformer is strictly reduced to $(L - m)$, bounding the final computational complexity to $\mathcal{O}((N_{\text{base}} + L - m)^2)$. This precise mask-guided pruning elevates the conditional token utilization from 70\% to over 90\%, ensuring that the computational budget is entirely dedicated to informative garment features.

\subsection{Multi-dimensional Try-on Reward (MTR)}
\label{subsec:mtr}

While the proposed ATP mechanism successfully mitigates memory explosion---making native 2K multi-garment training computationally feasible---merely scaling the resolution does not automatically guarantee photorealistic textures. In practical, standard Supervised Fine-Tuning  for virtual try-on encounters a fundamental quality bottleneck rooted in the inherent spectral bias of the MSE objective. The model tends to prioritize low-frequency structural alignment while critically neglecting the high-frequency details essential for fine-grained tactile realism. At native 2K resolution, this spectral bias leads to suboptimal material fidelity, causing complex textiles (e.g., wool, denim) to deteriorate into over-smoothed, ``plastic-like'' surfaces. To systematically restore these high-frequency components and ensure the rendering of authentic fabrics, we formulate the Multi-dimensional Try-on Reward (MTR), a post-training preference alignment pipeline that explicitly refines tactile realism via dual-dimensional objectives. Theoretically, MTR inherits the concepts of Direct Preference Optimization~\cite{rafailov2023direct} (DPO), fundamentally bypassing the computationally expensive policy sampling required by traditional reinforcement learning algorithms like PPO~\cite{schulman2017proximal}.

\subsubsection{Cloth Distribution Reward (CDR)}

The objective of CDR is to leverage the distributional margin between high-fidelity texture patterns and the over-smoothed artifacts inherent in the SFT stage. By shifting the generative trajectory toward the high-quality data manifold, CDR implicitly penalizes the low-pass filtering effect. 

Specifically, we curate a comparative subset based on the physical material fidelity of generated results. At each noisy timestep $t$, a preference pair $(x_t^w, x_t^l)$ is constructed by diffusing a high-quality (chosen) sample $x_0^w$ and a low-quality (rejected) sample $x_0^l$ to the same noise level. The alignment loss is formulated as:
\begin{equation}
    \mathcal{L}^{\text{CDR}}_t = - \mathbb{E}_{t, x_t^w, x_t^l} \left[ \log \sigma \left( \mathbf{R}_{\theta}(x_t^w, \mathcal{C}, \mathcal{P}, v_{\text{GT}}^w) - \mathbf{R}_{\theta}(x_t^l, \mathcal{C}, \mathcal{P}, v_{\text{GT}}^l) \right) \right],
\end{equation}

\noindent where $\sigma$ denotes the sigmoid function, $\mathcal{P}$ is the styling prompt, and $\mathcal{C}$ is the reference garment set. To avoid training an explicit reward model, the implicit reward $\mathbf{R}_{\theta}$ is defined directly via the velocity-field prediction errors between the trainable model $v_\theta$ and the frozen reference model $v_{\text{ref}}$:

\begin{equation}
    \mathbf{R}_{\theta}(x_t^*, \mathcal{C}, \mathcal{P}) = \beta \left( \| v_{\text{ref}}(x_t^*, \mathcal{C}, \mathcal{P}) - v_{\text{GT}}^* \|_2^2 - \| v_{\theta}(x_t^*, \mathcal{C}, \mathcal{P}) - v_{\text{GT}}^* \|_2^2 \right),
\end{equation}

\noindent where $\beta$ is a scaling hyper-parameter controlling the reward margin. $v_{\theta}$ and $v_{\text{ref}}$ denote the velocity fields predicted at timestep $t$ by the trainable model and the frozen reference model, respectively. $v_{\text{GT}}$ represents the corresponding ground-truth velocity target analytically computed for the current timestep. $*$ could be $w$ or $l$, which is the chosen or rejected samples at the same time.

\subsubsection{Semantic Guidance Reward (SGR)}

While the Cloth Distribution Reward (CDR) successfully calibrates the velocity field to restore fine-grained fabric textures, relying exclusively on visual preference optimization suffers from sluggish convergence. We attribute this bottleneck to the rigid image-text prior established during the prolonged SFT phase. Since the underlying foundation model strongly couples visual distributions with textual prompts, optimizing the visual branch in isolation creates significant optimization inertia. To explicitly break this barrier and accelerate multimodal alignment, we introduce a complementary textual objective inspired by Semantic Relative Preference Optimization~\cite{shen2025srpo}.

SGR widens the semantic similarity gap between the generated image and high-quality material prompts versus low-quality ones. First, given the clothing-agnostic model $I_{\text{agnostic}}$, the visual conditions $\mathcal{C}$, and the baseline prompt $\mathcal{P}$, the diffusion backbone predicts the velocity field $v_{\theta}(x_t, \mathcal{C}, \mathcal{P})$ at the noisy timestep $t$. Utilizing single-step denoising approximation, we project the current latent state directly to the pristine $t=0$ space, obtaining the native 2K spatial observation $\hat{I}_0$. To establish explicit contrastive semantic guidance, we construct paired semantic anchors: a positive prompt $\mathcal{P}_{\text{pos}}$ (e.g., highly detailed, authentic material descriptions) and a negative prompt $\mathcal{P}_{\text{neg}}$ (e.g., over-smoothed or plastic-like instructions). A frozen multi-modal reward model (e.g., CLIP~\cite{radford2021clip}) then computes their respective cosine similarities with the generated image at each timestep $t$, denoted as $\mathbf{S}_t^{\text{pos}}$ and $\mathbf{S}_t^{\text{neg}}$.

\begin{equation}
    \mathbf{S}_t^{\text{pos}} = \cos(f_{\text{img}}(\hat{I}_0), f_{\text{txt}}(\mathcal{P}_{\text{pos}})), \quad \mathbf{S}_t^{\text{neg}} = \cos(f_{\text{img}}(\hat{I}_0), f_{\text{txt}}(\mathcal{P}_{\text{neg}})),
\end{equation}
\noindent where $f_{\text{img}}$ and $f_{\text{txt}}$ denote the vision and text encoders, respectively. The Semantic Guidance Loss is formulated using a standard margin-based ReLU activation:
\begin{equation}
    \mathcal{L}^{\text{SGR}}_t = \text{ReLU} \left( \tau - \left[ (1+\omega) \mathbf{S}_t^{\text{pos}} -  \mathbf{S}_t^{\text{neg}} \right] \right),
\end{equation}
\noindent where $\omega > 0$ acts as the relative CFG scaling coefficient, and $\tau$ is the predefined semantic margin. 

A well-documented vulnerability of purely semantic-driven optimization is ``reward hacking'', where the model uncontrollably over-optimizes for textual alignment at the severe expense of structural integrity (e.g., inducing chromatic shifts or topological fragmentation). To effectively circumvent this degenerate behavior, we do not rely on SGR in isolation. Instead, our complete Multi-dimensional Try-on Reward (MTR) pipeline synergizes both dimensional objectives:
\begin{equation}
    \mathcal{L}^{\text{MTR}}_t = \alpha_t \mathcal{L}^{\text{CDR}}_t + \lambda \mathcal{L}^{\text{SGR}}_t,
\end{equation}
\noindent where $\lambda$ is a balancing coefficient, and $\alpha_t$ is a timestep-dependent weighting function designed to dynamically dampen the CDR gradient at extreme noise levels. 

This dual-dimensional balancing mechanism acts as a robust defense against reward hacking. The CDR implicitly anchors the generative trajectory within the physically correct high-frequency data distribution, preventing structural collapse, while the SGR provides explicitly accelerated semantic guidance for tactile restoration. Together, they enable the model to simultaneously fine-tune its outputs from both visual and textual perspectives.

\section{Experiment}

\subsection{Datasets}
\label{subsec:datasets}

To support native 2K multi-garment synthesis and empower mask-free training, we curate the WearWow-2K dataset, comprising approximately 100,000 training triplets and 2,000 test triplets at a pristine 2048 $\times$ 1536 resolution. To guarantee extreme visual quality and complex spatial diversity, we employ a hybrid curation strategy. Unlike existing single-item benchmarks, WearWow-2K features sophisticated layer-aware styling combinations ranging from 1 to 6 concurrent items per subject. Its highly diverse sartorial ontology extends beyond basic apparel to include footwear and fine-grained accessories (e.g., earrings), providing the physically correct spatial interactions necessary for multi-condition learning. Detailed data distributions and qualitative visual examples are deferred to the supplementary material.

\subsection{Implementation Details}
\label{subsec:implementation}

We build the WearWow generative backbone upon the pre-trained Qwen-Image-Edit foundation model, extracting hierarchical features from reference garments and textual prompts via a Qwen2.5VL~\cite{yang2025qwen3} vision-language encoder coupled with our Adaptive 2D Token Packing (ATP) module. During the initial Supervised Fine-Tuning (SFT) phase at native 2K resolution ($2048 \times 1536$).
And a global batch size of $8$ for $80$K steps. Following SFT convergence, we execute the Multi-dimensional Try-on Reward (MTR) alignment to explicitly restore tactile realism. The SGR text-image similarities are computed using a frozen HPSv2~\cite{wu2023human}. MTR fine-tuning runs for an additional 800 steps at a reduced learning rate of $1 \times 10^{-5}$, with hyperparameters empirically set to CFG scale $\omega \in [0.1, 0.6]$ (sampled as the timesteps changing), safety margin $\tau = 0.5$, CDR scale $\beta = 0.5$, and balancing weight $\lambda = 0.65$. All training is conducted in \textit{bf16} on 8 $\times$ NVIDIA H20 GPUs.

\begin{figure}[!t]
  \centering
  \includegraphics[width=1.0\textwidth]{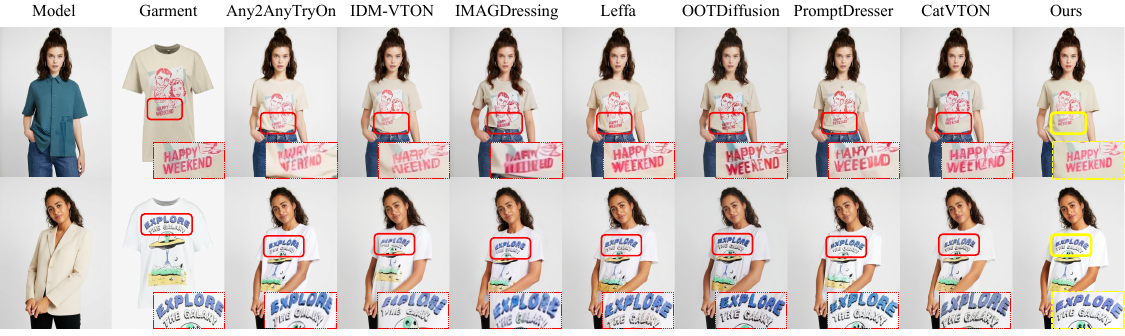}
  \caption{Qualitative comparisons on VITON-HD and DressCode in the single try-on. Compared with other methods, WearWow produces more texture-consistent images.}
  \label{fig:single}

\end{figure}

\subsection{Baselines and Experimental Setup}
\label{subsec:setup}

To comprehensively evaluate the robustness and scalability of WearWow, we design a dual-track evaluation protocol: a standard single-garment setting and an ultra-high-resolution multi-garment setting.

\noindent\textbf{Single-Garment Setup.} 
To ensure a rigorous and comprehensive evaluation, we conduct our single-garment experiments on a unified test dataset that includes the widely adopted VITON-HD~\cite{choi2021vitonhd} and DressCode~\cite{morelli2022dresscode} benchmarks. Since existing open-source baselines strictly limit their scope to this single-item paradigm, we evaluate this specific task at the standard $1024 \times 768$ resolution. We benchmark WearWow against a comprehensive suite of state-of-the-art diffusion-based models, including Any2AnyTryOn~\cite{guo2025any2anytryon}, IDM-VTON~\cite{choi2024idmvton}, IMAGDressing~\cite{shen2024IMAGDressing-v1}, Leffa~\cite{zhou2024leffa}, OOTDiffusion~\cite{xu2024ootdiffusion}, PromptDresser~\cite{kim2024promptdresserimprovingqualitycontrollability}, and CatVTON~\cite{chong2024catvtonconcatenationneedvirtual}.

\noindent\textbf{Multi-Garment Setup.} 
To evaluate the extreme combinatorial complexity of multi-garment synthesis, we utilize our proposed WearWow-2K dataset at its native $2048 \times 1536$ resolution. Because traditional open-source try-on models mathematically collapse (OOM) under multi-condition 2K inputs, we benchmark WearWow against the most powerful recent generative foundation models and commercial APIs. The baselines include local multi-modal models (OminiTry ~\cite{feng2025omnitry}, QWen-image~\cite{wu2025qwen}) and top-tier closed-source APIs (Nano Banana Pro~\cite{saharia2022imagen}, KeLingv3~\cite{team2025kling}, Seedream4.0~\cite{seedream2025seedream} and GPT-image 1.5~\cite{betker2023dalle3}).

\noindent\textbf{Evaluation Metrics.} 
We employ a robust set of quantitative metrics to assess generative quality and structural alignment. To explicitly quantify our core contribution in restoring high-frequency tactile realism and perceptual quality, we prioritize the Learned Perceptual Image Patch Similarity (LPIPS~\cite{zhang2018lpips}) and Fréchet Inception Distance (FID~\cite{Seitzer2020FID}). We also report Kernel Inception Distance (KID~\cite{bińkowski2021kid}) for unbiased distribution measurement and Structural Similarity Index (SSIM~\cite{wang2004ssim}) for spatial alignment. For ablation study, CLIP score is also computed by measuring the similarity between textual description and the result. 

\noindent\textbf{Human Evaluation (HE).} 
To complement quantitative metrics that occasionally miss micro-level perceptual nuances at 2K resolution, we conduct a rigorous Human Evaluation with 100 diverse participants. Evaluators blindly score randomized outputs from WearWow and baselines across three dimensions: (1) Material Fidelity (tactile realism without plastic artifacts), (2) Spatial Coherence (correct multi-garment occlusion), and (3) Overall Photorealism.The averaged Human Evaluation (HE) scores are reported to provide a definitive subjective benchmark.

\begin{figure}[!t]
  \centering
  \includegraphics[width=1.0\textwidth]{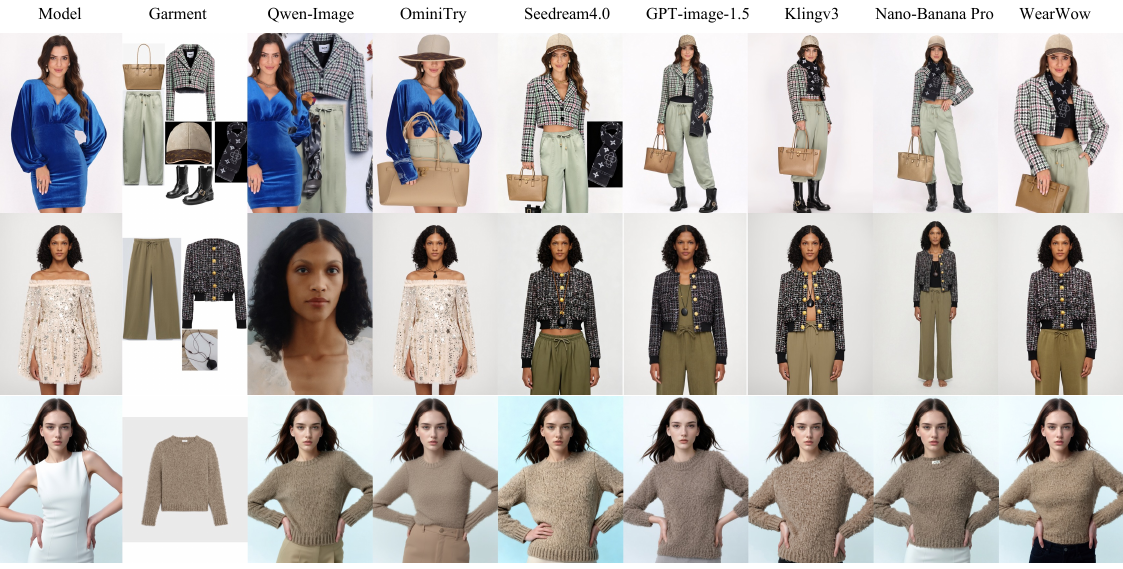}
  \caption{Qualitative comparisons on WearWow-2k,showcasing our superior photorealism on complex multi-item combinations and challenging woolen textures.}
  \label{fig:multi}

\end{figure}

\subsection{Quantitative Analysis}
\label{subsec:quantitative}

\noindent\textbf{Single-Garment Evaluation.} 
As reported in Table~\ref{tab:openbench}, WearWow demonstrates highly competitive performance in perceptual realism. Specifically, WearWow achieves strong results across distribution and perceptual metrics, showing notable improvements over recent baselines like IDM-VTON~\cite{choi2024idmvton} and Leffa~\cite{zhou2024leffa} in terms of FID and LPIPS. It is worth noting that while CatVTON~\cite{chong2024catvtonconcatenationneedvirtual} achieves a marginally higher SSIM, this metric heavily relies on pixel-wise mean error, which can inherently favor models that produce overly smoothed, low-frequency outputs. In contrast, WearWow's improvements in FID, KID, and LPIPS indicate that our MTR pipeline effectively restores high-frequency tactile realism. Furthermore, our Human Evaluation (HE) supports this alignment with human visual preferences; evaluators generally favored WearWow in terms of material fidelity and overall photorealism.

\noindent\textbf{Multi-Garment Evaluation.} 
The quantitative advantages of WearWow are further evident in the native 2K multi-garment regime (Table~\ref{tab:multi}). Commercial APIs like Nano Banana Pro~\cite{saharia2022banana} and KeLingv3~\cite{team2025kling}, despite their large parameter counts, often encounter difficulties with strict spatial control and structural consistency, reflecting in their FID and KID scores. WearWow effectively addresses these challenges, achieving consistent improvements across key metrics (FID, KID, SSIM, and LPIPS). This stable performance indicates that our Adaptive 2D Token Packing successfully mitigates the spatial interference of multi-garment conditions, while MTR maintains the generative quality at extreme spatial scales. Consistent with the objective metrics, our Human Evaluation (HE) shows a clear user preference for WearWow, particularly highlighting its spatial coherence and mask-free compositional stability compared to closed-source foundation models.

\begin{table*}[t]
\centering
\caption{Quantitative comparisons on the VITON-HD and DressCode dataset.}
\setlength{\tabcolsep}{3mm}

\begin{tabular}{l c c c c c}
    \toprule
    Method & SSIM $\uparrow$ & FID $\downarrow$ & LPIPS $\downarrow$ & KID $\downarrow$ & HE $\uparrow$ \\
    \midrule
    Any2AnyTryOn~\cite{guo2025any2anytryon}  & 0.8460 & 6.7285 & 0.1292 & 1.4906 & 0.1138 \\
    IDM-VTON~\cite{choi2024idmvton}      & 0.8779 & 5.9893 & 0.0720 & 1.7433 & 0.0863 \\
    IMAGDressing~\cite{shen2024IMAGDressing-v1}  & 0.8298 & 9.0388 & 0.1120 & 2.8076 & 0.0379 \\
    Leffa~\cite{zhou2024leffa}         & 0.8582 & 5.9421 & 0.0825 & 0.9899 & 0.1389 \\
    OOTDiffusion~\cite{xu2024ootdiffusion}  & 0.8565 & 5.9640 & 0.0782 & 0.9249 & 0.0505 \\
    PromptDresser~\cite{kim2024promptdresserimprovingqualitycontrollability} & 0.8513 & 6.5618 & 0.0973 & 1.3420 & 0.0905 \\
    CatVTON~\cite{chong2024catvtonconcatenationneedvirtual}       & \textbf{0.8950} & 4.5698 & 0.0746 & 1.0298 & 0.0737 \\
\midrule
\textbf{WearWow}        & 0.8815 & \textbf{4.2475} & \textbf{0.0688} & \textbf{0.6503} & \textbf{0.4084} \\
    \bottomrule

\end{tabular}
\label{tab:openbench}
\end{table*}

\begin{table*}[t]
\setlength{\tabcolsep}{2.6mm}

\centering
\caption{Quantitative comparisons on WearWow-2K .}

\begin{tabular}[t]{l c c c c c}
  \toprule
  Method & FID $\downarrow$ & KID $\downarrow$ & LPIPS $\downarrow$ & SSIM $\uparrow$ & HE $\uparrow$ \\
  \midrule
    OminiTry~\cite{feng2025omnitry}    & 12.4164 & 2.0228 & 0.1444 & 0.8002 & 0.0257 \\
    QWen-image~\cite{wu2025qwen}  & 14.3538 & 2.3196 & 0.2175 & 0.7516 & 0.0286 \\
    Seedream4.0~\cite{seedream2025seedream}                    & 14.3389 & 4.7620 & 0.2175 & 0.7699 & 0.1514 \\
    Nano Banana Pro~\cite{saharia2022banana}    & 8.1823  & 0.8285 & 0.1114 & 0.8215 & 0.2229 \\
    KeLingv3~\cite{team2025kling}           & 8.9624  & 1.0006 & 0.1108 & 0.8255 & 0.1314 \\
    GPT-image 1.5~\cite{betker2023dalle3}      & 10.7218 & 1.3345 & 0.1724 & 0.7705 & 0.1857 \\
    \midrule
    \textbf{WearWow (ours)}          & \textbf{7.5400} & \textbf{0.2643} & \textbf{0.0773} & \textbf{0.8461} & \textbf{0.2543} \\
\bottomrule
\end{tabular}
\label{tab:multi}

\end{table*}

\subsection{Qualitative Analysis}
\label{subsec:qualitative}

\noindent\textbf{Single-Garment Tactile Realism and Consistency.} 
As illustrated in Figure~\ref{fig:single}, existing baselines struggle severely when rendering garments with complex physical properties. Models like IDM-VTON~\cite{choi2024idmvton} and CatVTON~\cite{chong2024catvtonconcatenationneedvirtual} tend to act as low-pass filters, degrading intricate materials into synthetic, plastic-like flat surfaces. Conversely, empowered by the dual-dimensional MTR alignment, WearWow exhibits superior capability in restoring high-frequency details. Notably, when processing challenging fabrics like rugged denim or complex texture garment , our framework demonstrates exceptional stability and consistency. 

\noindent\textbf{Multi-Garment Stability and Material Fidelity.} 
Figure~\ref{fig:multi} demonstrates the mask-free joint synthesis capabilities on WearWow-2K. When prompted to simultaneously wear multiple items (e.g., inner-shirts, thick outerwear, denim pants, and fine accessories), recent open-source models and baseline APIs often exhibit severe compositional instability. They either hallucinate bizarre merged textures (feature leakage), entirely ignore specific clothing items, or suffer from catastrophic texture degradation at native 2K resolution. In stark contrast, WearWow operates with impeccable generative stability. Driven by the layer-aware annotations and the ATP module's strict 2D positional preservation, WearWow accurately renders correct depth orderings (e.g., a jacket explicitly open over a tucked-in shirt) with flawless occlusion boundaries, while simultaneously maintaining extreme material fidelity across all concurrent garments.


\subsection{Ablation study}

\label{subsec:ablation}

\begin{figure}[!t]
  \centering
  \includegraphics[width=1.0\textwidth]{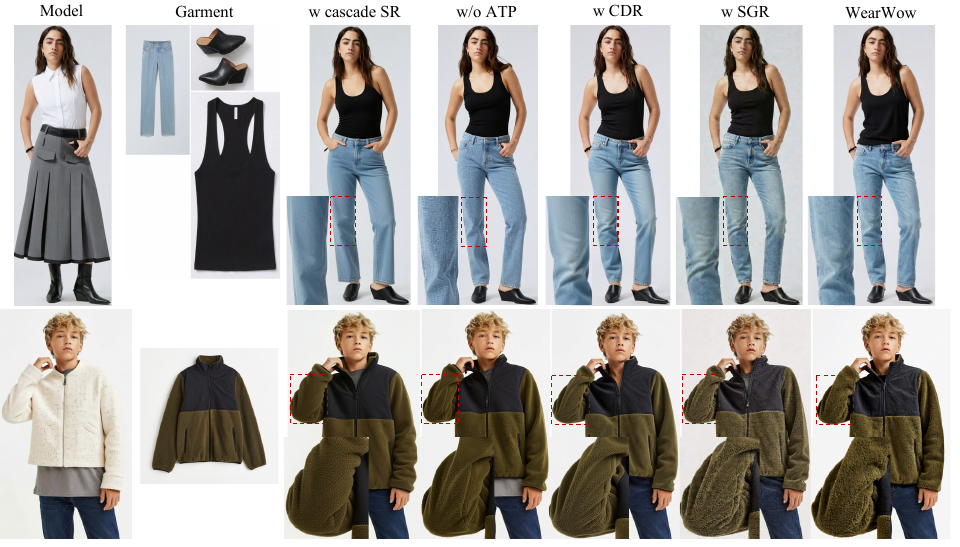}
  \caption{We visualize the impact of our proposed modules. our full MTR pipeline (\textbf{Ours}) robustly restores micro-level tactile realism without sacrificing structural integrity.}
  \label{fig:ablation}

\end{figure}



To rigorously validate our core contributions, we evaluate the spatial representation strategy (ATP) and preference alignment pipeline (MTR) against various baselines (Table~\ref{tab:ablation} and Figure~\ref{fig:ablation}). To validate ATP, we compare WearWow against two scaling paradigms. The \textbf{w/ Cascaded SR} variant processes inputs at a lower resolution before employing an external Super-Resolution model, which inevitably introduces structural hallucinations. Alternatively, maintaining native 2K training but naively downscaling reference items to satisfy memory constraints (\textbf{w/o ATP}) severely destroys high-frequency visual priors. Consequently, both approaches suffer from noticeable texture degradation and catastrophic over-smoothing, exhibiting inferior scores compared to our ATP-equipped model. This demonstrates that algorithmically packing and pruning background tokens is essential for preserving exact foreground resolution and spatial inductive biases.

\begin{wraptable}{r}{0.6\textwidth}
\vspace{-3mm}
\caption{ Comparison of our full WearWow framework against its ablative variants.
}
\begin{tabular}{l c c c}
  \toprule
  Method  & SSIM $\uparrow$ & LPIPS $\downarrow$ & CLIP Score $\uparrow$ \\
  \cmidrule(lr){1-1}\cmidrule(lr){2-4}
   w/ cascade SR   & 0.8502 & 0.1237 & 0.2651 \\
   w/o ATP      & 0.8468 & 0.0923 & 0.2654  \\
  \cmidrule(lr){1-1}\cmidrule(lr){2-4}
  w/ SGR      & 0.8599 & 0.0874 & 0.2678 \\
  w/ CDR      & 0.8596 & 0.0961 & 0.2664  \\
  \cmidrule(lr){1-1}\cmidrule(lr){2-4}
  \textbf{Ours} & \textbf{0.8677} & \textbf{0.0870} & \textbf{0.2690} \\
  \bottomrule
\end{tabular}
\vspace{-3mm}

\label{tab:ablation}
\end{wraptable}

Next, we dissect the MTR pipeline to validate the synergy between its physical and semantic components. Applying only the Cloth Distribution Reward (\textbf{w/ CDR}) anchors the generative trajectory to the data distribution, yielding slight visual improvements. However, its training convergence is notably slow, and it ultimately struggles to fully overcome the SFT optimization inertia, leaving the tactile realism of complex high-frequency textures insufficient. Conversely, relying solely on the Semantic Guidance Reward (\textbf{w/ SGR}) provides a strong textual push but exhibits critical vulnerabilities. Pure semantic guidance initially forces an over-alignment with textual prompts, introducing unnatural chromatic shifts and an excessively rough, frizzy appearance to the fabrics. More critically, as training progresses, it inevitably induces severe ``reward hacking''. By synergizing both objectives, our full MTR pipeline successfully resolves all aforementioned issues; CDR provides the essential physical anchor to ensure stability and prevent hacking, while SGR effectively accelerates convergence and restores flawless micro-level tactile realism.

\section{Conclusion}
\label{sec:conclusion}

In this paper, we introduced \textbf{WearWow}, an effective generative framework that scales virtual try-on to the native 2K multi-garment regime. To dismantle the prohibitive memory barriers associated with extreme-resolution condition injection, we proposed Adaptive 2D Token Packing . By exploiting the inherent spatial sparsity of garment images, ATP algorithmically packs and prunes heterogeneous visual tokens into a unified 2D layout, effectively mitigate the computational explosion while rigorously preserving essential 2D spatial priors. Furthermore, to address the inherent spectral bias and over-smoothing degradation of standard diffusion models, we formulated the Multi-dimensional Try-on Reward. By synergizing a Cloth Distribution Reward to anchor the physical distribution with a Semantic Guidance Reward  to provide explicit textual pushes, MTR aligns the generative trajectory to penalize plastic-like artifacts and robustly restore micro-level tactile realism. Supported by the  WearWow-2K dataset, our framework achieves both precise spatial composition and authentic fabric details and sets a state-of-the-art for 2K multi-garment virtual try-on. 

\noindent\textbf{Social Impact}: While WearWow significantly advances digital fashion, we acknowledge the inherent risks of ultra-high-fidelity human generation. We emphasize the critical need for robust deepfake detection algorithms and strictly regulated deployment to prevent the malicious misuse of generated identities.

\noindent\textbf{Limitations and Future Work}
While WearWow successfully scales multi-garment virtual try-on to native 2K resolution and establishes a new standard for tactile realism, it possesses certain limitations that present exciting avenues for future research. Handling extreme complex back-view occlusions remains a common challenge. Future work will explore integrating explicit 3D human body priors to guarantee rigorous, view-consistent multi-garment synthesis across arbitrary camera angles. Furthermore, building upon the robust spatial representation and tactile foundation established by WearWow, a natural and highly promising progression is extending this framework into the temporal domain. We aim to explore temporal consistency mechanisms for high-resolution video virtual try-on, coupled with dynamic garment relighting under complex environmental illuminations. 

\section*{Acknowledgements}

This work is supported by National Key Research and Development Program of China (2024YFE0203100), Scientific Research Innovation Capability Support Project for Young Faculty (No.ZYGXQNJSKYCXNLZCXM-I28), Shenzhen Science and Technology Program (Grant No.QNXMA20250701093544048) and General Embodied AI Center of Sun Yat-sen University.

%
%
\bibliographystyle{splncs04}
\bibliography{main}
\end{document}


\title{Supplementary for WearWow} 
\titlerunning{WearWow}

\author{Xujie Zhang\inst{1}\setcounter{footnote}{0}\thanks{Equal contribution.} \and
Runyan Du\inst{2}$^{\star}$ \and
Song Chang\inst{2} \and
Jiang Li\inst{2} \and
Dongliang Shao\inst{2} \and
Liping Wu\inst{2} \and
Luo Wei\inst{2} \and
Xiaochao Qu\inst{2} \and
Luoqi Liu\inst{2} \and
Xiaodan Liang\inst{1}\setcounter{footnote}{3}\thanks{Corresponding author.}}

\authorrunning{X. Zhang et al.}

\institute{Shenzhen Campus of Sun Yat-sen University \\
\email{zhangxj59@mail2.sysu.edu.cn, liangxd9@mail.sysu.edu.cn} \and
Meitu Lab \\
\email{\{dry, cs2, lj28, sdl, wlp1, lw20, qxc, llq5\}@meitu.com}}
\maketitle

In this supplementary material, we provide extended details regarding our Native 2K training dataset construction and the data synthesis pipeline (Sec.~\ref{sec:data} and Sec.~\ref{sec:extended_method}). Furthermore, we present a detailed computational analysis of our Adaptive Token Packing (ATP) module (Sec.~\ref{sec:comp_analysis}), provide the code for our method (Sec.~\ref{sec:atp_algo}), and include comprehensive qualitative results, covering in-the-wild generalization, visual comparisons, and failure cases (Sec.~\ref{sec:AR}).

\section{Native 2K Multi-Garment Data Construction}
\label{sec:data}
The severe scarcity of ultra-high-resolution, multi-garment paired data remains the primary bottleneck in digital fashion synthesis. To establish a robust structural foundation for native 2K generation, we curate an extreme-quality text-image training triplet dataset: $\langle I_{\text{agnostic}}, \mathcal{C}, I_{\text{target}} \rangle$, where $I_{\text{agnostic}}$ is the clothing-agnostic input $I_{\text{agnostic}}$ (\textbf{the fake model}), $\mathcal{C}$ is the cloth, $I_{\text{target}}$ is the groundtruth. Crucially, to empower the network with mask-free generative capabilities and absolute textural authenticity, we adopt a ``Real Ground-Truth, Synthesized Fake Model'' paradigm.

\noindent\textbf{Fake Model Synthesis and Quality Control.}
Unlike previous methods that rely on synthetic pseudo-ground-truths, we strictly utilize authentic 2K photographs of models wearing multiple complex garments as our absolute ground-truth $I_{\text{target}}$. This guarantees the provision of flawless high-frequency texture gradients and physically correct multi-garment spatial interactions. To train the network to generate these garments without manual masks, we inversely synthesize the corresponding fake model. We engineer a hybrid erasing pipeline synergizing our custom-trained try-off model, Nano Banana 2, and auxiliary try-on pipelines to systematically strip the target garments from $I_{\text{target}}$. 

Since the synthesis of $I_{\text{agnostic}}$ inevitably introduces occasional structural artifacts that could poison the generation target, we subject the generated triplets to an extremely strict manual screening protocol. The human filtering process rigorously evaluates three dimensions:
\begin{itemize}
    \item \textbf{Identity and Pose Preservation:} Ensuring the facial ID, background, body shape, and skeletal pose of the synthesized model strictly align with the original ground truth.
    \item \textbf{Synthesis Realism:} Verifying the structure of the fake model, ensuring no visible erasing traces, unnatural blending artifacts, or distortions exist.
    \item \textbf{Zero Information Leakage:} Strictly confirming that the fake model does not leak ground-truth garment priors. For instance, the silhouettes, hemlines, or cast shadows of the original garments must be entirely eliminated so the network cannot take ``shortcuts'' during the reconstruction process.
\end{itemize}
Through this aggressive curation, we yield a highly refined dataset comprising approximately 100,000 pristine native 2K triplets.

\noindent\textbf{Training Strategy.}
To prevent overfitting and enhance the model's generalizability across diverse real-world scenarios, we introduce several dynamic augmentation strategies during the training phase. First, we apply multi-scale resolution augmentation, randomly scaling the longest edge of the images between 1024 and 2048 pixels across varying aspect ratios (1:1, 2:3, 3:4, and 9:16). Second, during the Adaptive 2D Token Packing process, the relative area proportions of different garment categories are dynamically perturbed within a defined range. This spatial augmentation forces the network to learn scale-invariant reasoning across varying positional layouts. Finally, we implement a probabilistic multi-task optimization scheme. The forward pass randomly alternates between try-on synthesis and reference garment Tryoff, calculating respective losses with tailored weights to guide a balanced gradient backpropagation.

\noindent\textbf{Multimodal Semantic Annotation.}
To enforce rigorous text-vision alignment, we deploy a state-of-the-art Large Vision-Language Model to extract hierarchical semantics of the garment. From the condition set $\mathcal{C}$, we extract intrinsic physical descriptors (e.g., category, silhouette, fabric material). Most importantly, we place a strong emphasis on addressing the complex layering problems inherent in multi-garment scenarios. From $I_{\text{target}}$, the VLM meticulously parses inter-garment spatial interactions, specifically identifying depth ordering, occlusion relationships, and styling states (e.g., outerwear worn over an inner-wear, a shirt tucked into denim pants). These critical layering dynamics are aggregated into a deterministic directive prompt $\mathcal{P}$ (e.g., ``Dress the model in a coarse wool coat over a silk inner-shirt, and 3/4 denim pants. The coat is unbuttoned revealing the inner-shirt.''). This explicit layer-aware annotation ensures the network does not merely paste garments, but physically composes them according to text-guided depth topologies.

\section{Algorithm for Adaptive 2D Token Packing}
\label{sec:atp_algo}
To provide a comprehensive understanding of the spatial recombination process, we detail the Adaptive 2D Token Packing mechanism in Algorithm~\ref{alg:atp}. The algorithm dynamically scales and packs heterogeneous reference garments onto a unified canvas to inject scale-invariant spatial augmentation. Subsequently, a rigorous mask-guided pruning stage is executed to eliminate uninformative padding, strictly bounding the conditional sequence length to $L - m$ while seamlessly preserving essential 2D positional priors.
\begin{algorithm}[htbp]
\caption{Adaptive 2D Token Packing (ATP) and Mask-Guided Pruning}
\label{alg:atp}
\textbf{Input:} Reference garment set $\mathcal{C} = \{c_1, c_2, \dots, c_K\}$, Target canvas size $H \times W$ (Yielding token capacity $L$), Spatial augmentation range $[\alpha_{\min}, \alpha_{\max}]$.\\
\textbf{Output:} Pruned condition sequence $S_{\text{pruned}}$, Updated 2D Positional Encodings $P_{\text{2D}}$.
\begin{algorithmic}[1]
\State Initialize empty canvas $V_{\text{canvas}}, V_{\text{canvas-tmp}} \gets \mathbf{0}^{H \times W \times 3}$
\State Initialize binary mask canvas $M_{\text{canvas}},M_{\text{canvas-tmp}} \gets \mathbf{0}^{H \times W}$
\State Initialize 2D bounding box allocator $\mathcal{A}(H, W)$
\State Initialize Spatial utilization $U \gets 0$
\State Initialize Maximum number of trial $T_{max} \gets 30$
\Statex \Comment{\textit{Stage 1: Dynamic Spatial Augmentation and 2D Packing}}

\For{$trial = 1$ \textbf{to} $T_{max}$}
    \State Shuffle($\mathcal{C}$) \Comment{Each trial uses a random order}
    \For{each garment $c_k \in \mathcal{C}$}
        \State Extract tight bounding box $B_k$ and valid pixel mask $m_k$ from $c_k$
        \State Sample dynamic spatial scaling factor $\alpha \sim U(\alpha_{\min}, \alpha_{\max})$
        \State Scale $B_k$ and $m_k$ by factor $\alpha$ \Comment{Injects scale-invariant prior}
        \State $(x_k, y_k) \gets \mathcal{A}.\text{find\_position}(B_k)$ \Comment{Maximal Rectangles Algorithms}
        \State $V_{\text{canvas-tmp}}[x_k, y_k] \gets c_k[B_k]$ \Comment{Paste garment pixels}
        \State $M_{\text{canvas-tmp}}[x_k, y_k] \gets m_k$ \Comment{Update global valid mask}
    \EndFor
    \If{$\frac{\sum^K_{k=1}m_k}{H \times W} \geq U$} 
    \Comment{Determine of the current highest utilization rate}
        \State $V_{\text{canvas}} \gets V_{\text{canvas-tmp}}$
        \State $M_{\text{canvas}} \gets M_{\text{canvas-tmp}}$
        \State $U \gets \frac{\sum^K_{k=1}m_k}{H \times W}$
    \EndIf
\EndFor

\Statex \Comment{\textit{Stage 2: Tokenization and 2D Positional Encoding Extraction}}
\State $S_{\text{full}} \gets \text{Patchify}(V_{\text{canvas}})$ \Comment{Sequence length: $L$}
\State $P_{\text{full}} \gets \text{Extract\_2D\_PE}(H, W)$ \Comment{Preserve unified 2D coordinates}
\State $M_{\text{tokens}} \gets \text{Patchify\_Mask}(M_{\text{canvas}})$ \Comment{Token-level binary validity}
\Statex \Comment{\textit{Stage 3: Sparsity-Driven Token Pruning}}
\State Initialize $S_{\text{pruned}} \gets [\,]$, $P_{\text{2D}} \gets [\,]$
\For{$i = 1$ \textbf{to} $L$}
    \If{$M_{\text{tokens}}[i] == 1$} \Comment{Discard $m$ uninformative background tokens}
        \State $S_{\text{pruned}}.\text{append}(S_{\text{full}}[i])$
        \State $P_{\text{2D}}.\text{append}(P_{\text{full}}[i])$
    \EndIf
\EndFor
\State \textbf{return} $S_{\text{pruned}}, P_{\text{2D}}$ \Comment{Final effective sequence length: $L - m$}
\end{algorithmic}
\end{algorithm}

\section{Extended Methodology Details}
\label{sec:extended_method}

\subsection{Extended Data Synthesis \& Filtering Pipeline}
As discussed in the main text, constructing paired data is challenging. To supplement paired data for complex multi-garment interactions, we utilize commercial APIs (e.g., Nano Pro) via three specific strategies: (1) augmenting single-garment pairs with new items (e.g., hats) to create multi-garment data, (2) swapping garments to increase cross-category diversity, and (3) erasing or replacing items on raw shop-model pairs. To strictly prevent pose misalignment, facial alterations, or data leakage during this process, all synthesized data undergoes an automated VLM-filtering pipeline, followed by the rigorous human screening detailed in Sec.~\ref{sec:data}.

\subsection{Auxiliary Try-off Supervision}
During the development of our pipeline, we observed that the final output fidelity highly correlates with Ground Truth (GT) quality. Because some synthesized GTs (agnostic models) may be sub-optimal while their corresponding flat reference garments are ultra-high-definition, we introduce an auxiliary ``try-off'' supervision task during training. This mechanism elegantly leverages the high-quality reference garment features to optimize gradient updates, significantly elevating the final 2K generation quality and preventing the network from inheriting artifacts from the synthesized conditions.

\section{Computational Analysis of Adaptive Token Packing}
\label{sec:comp_analysis}

A core motivation of our Adaptive Token Packing (ATP) module is averting the computational explosion inherent in multi-condition high-resolution training. Standard baselines naively concatenate image tokens, which scales linearly with the number of conditions. 

As shown in Tab.~\ref{tab:comp_cost}, we evaluate the computational overhead at a 1536$\times$1152 resolution on a single H20 GPU. Processing just 2 garments with the naive concatenation baseline requires 62.6 GB VRAM and a sluggish 66.7 seconds latency. Attempting to scale this naive approach to 6 items inevitably triggers an Out-Of-Memory (OOM) error. 

In stark contrast, our ATP dynamically packs and prunes tokens onto a unified canvas, effectively decoupling the sequence length from the garment count $K$. Consequently, for the same 2-garment setting, WearWow slashes latency by over 50\% (to 32.2s) and stabilizes VRAM at 61.4 GB. This efficiency scales robustly, maintaining bounded overhead even when rendering the upper limit of 6 concurrent garments.

\begin{table}[h]
\centering
\caption{\textbf{Computational Efficiency Comparison.} Evaluated at 1536$\times$1152 resolution on an H20 GPU. ATP effectively decouples sequence length from garment count $K$.}
\label{tab:comp_cost}
\resizebox{0.95\linewidth}{!}{
\begin{tabular}{lcccc}
\toprule
\multirow{2}{*}{Method} & \multicolumn{2}{c}{$K=2$ Garments} & \multicolumn{2}{c}{$K=6$ Garments} \\
\cmidrule(lr){2-3} \cmidrule(lr){4-5}
 & VRAM (GB) $\downarrow$ & Latency (s) $\downarrow$ & VRAM (GB) $\downarrow$ & Latency (s) $\downarrow$ \\
\midrule
Naive Concatenation & 62.6 & 66.7 & OOM & - \\
\textbf{WearWow (Ours w/ ATP)} & \textbf{61.4} & \textbf{32.2} & \textbf{Bounded} & \textbf{Stable} \\
\bottomrule
\end{tabular}
}
\end{table}

\section{Additional Qualitative Results and Limitations}
\label{sec:AR}

\subsection{Comparison with Cascaded SR Models}
Recent paradigms have explored 1K generation followed by Super-Resolution (SR) cascaded pipelines. While inspiring, cascaded upsampling inevitably hallucinates high-frequency artifacts and alters authentic textures of the garments. Conversely, our Native 2K training avoids this spectral bias, preserving the true physical properties of the fabrics. 

\subsection{In-the-wild Generalization and Mask Validations}
\begin{figure}[h] 
  \centering
  \includegraphics[width=\textwidth]{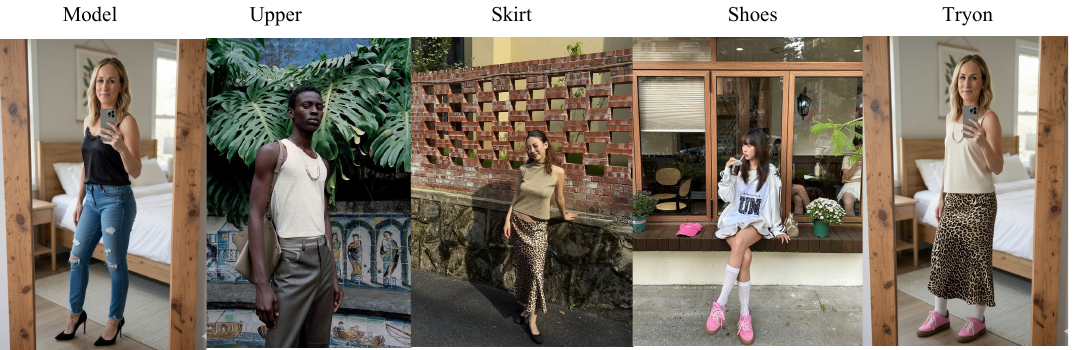} 
  \vspace{-2mm}
  \caption{\textbf{In-the-wild generalization.} WearWow robustly handles reference garments with complex backgrounds, demonstrating the scene-agnostic capability of our mask extraction and synthesis pipeline.}
  \label{fig:wild_supp}
\end{figure}

The perceived prevalence of studio shots in typical high-resolution datasets stems from their natural prioritization of close-ups to highlight fabric details. However, WearWow is entirely scene-agnostic. Fig.~\ref{fig:wild_supp} presents an in-the-wild result where all reference garments feature complex backgrounds. This high-quality synthesis implicitly validates the robustness of our internally fine-tuned segmentation network in extracting accurate valid-region masks.

\begin{figure}[h] 
  \centering
  \includegraphics[width=\linewidth]{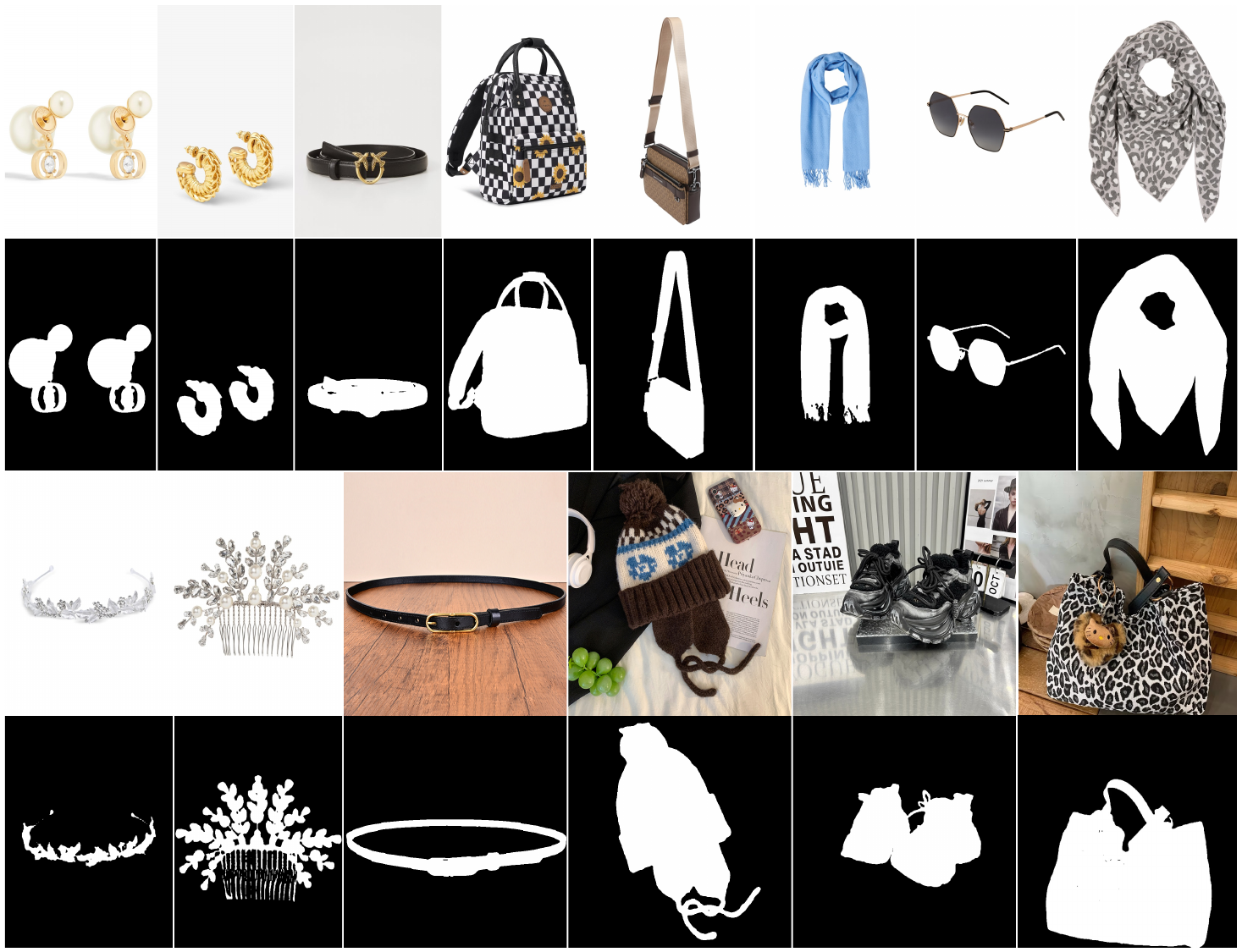}
   \caption{\textbf{Mask Visualizations.} Explicit visualizations of the valid-region masks extracted by our segmentation network.}
   \label{fig:masks}
\end{figure}

\subsection{Upper Bound and Failure Cases}
\begin{figure}[h] 
  \centering
  \includegraphics[width=\textwidth]{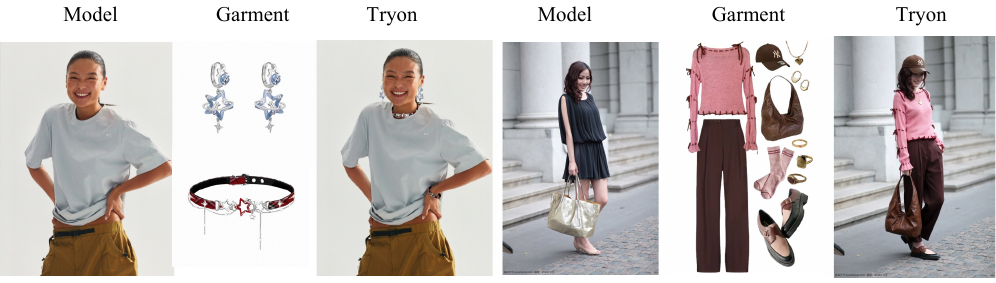}

  \caption{\textbf{Failure Cases.} (Left) Loss of fine details in micro-accessories. (Right) Item omission due to spatial constraints or capacity limits.}
  \label{fig:bad_supp}
\end{figure}

As demonstrated, WearWow robustly supports up to 6 concurrent garments via the unified canvas. However, we observe two primary limitations (illustrated in Fig.~\ref{fig:bad_supp}):\textbf{Micro-accessories:} Items such as delicate necklaces or thin rings may lose fine details. This is primarily due to their extreme scarcity and pixel-level sparsity in our current Native 2K training data. \textbf{Item Omission:} Omission occurs either when conditioned items physically fall outside the target's visible frame (e.g., trying to generate boots on a half-body cropped image), or when pushing the system beyond its practical 6-item capacity.

We plan to continuously expand our dataset diversity to mitigate these corner cases in future iterations.

\section{Additional Results}\label{sec:AR}

\begin{figure}[!t]
  \centering
  \includegraphics[width=1.0\textwidth]{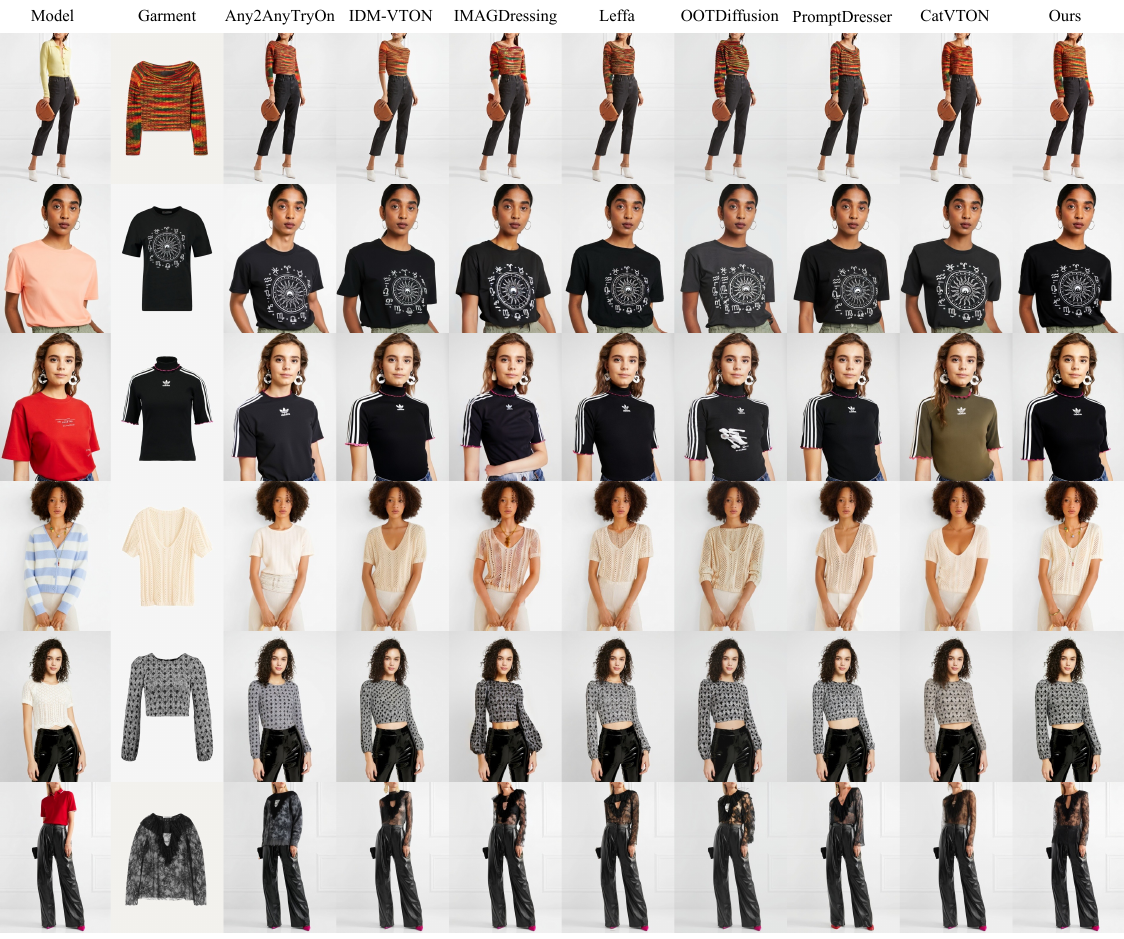}
  \caption{Qualitative comparisons on open benchmark}
  \label{fig:supp2}
\end{figure}

\begin{figure}[!t]
  \centering
  \includegraphics[width=1.0\textwidth]{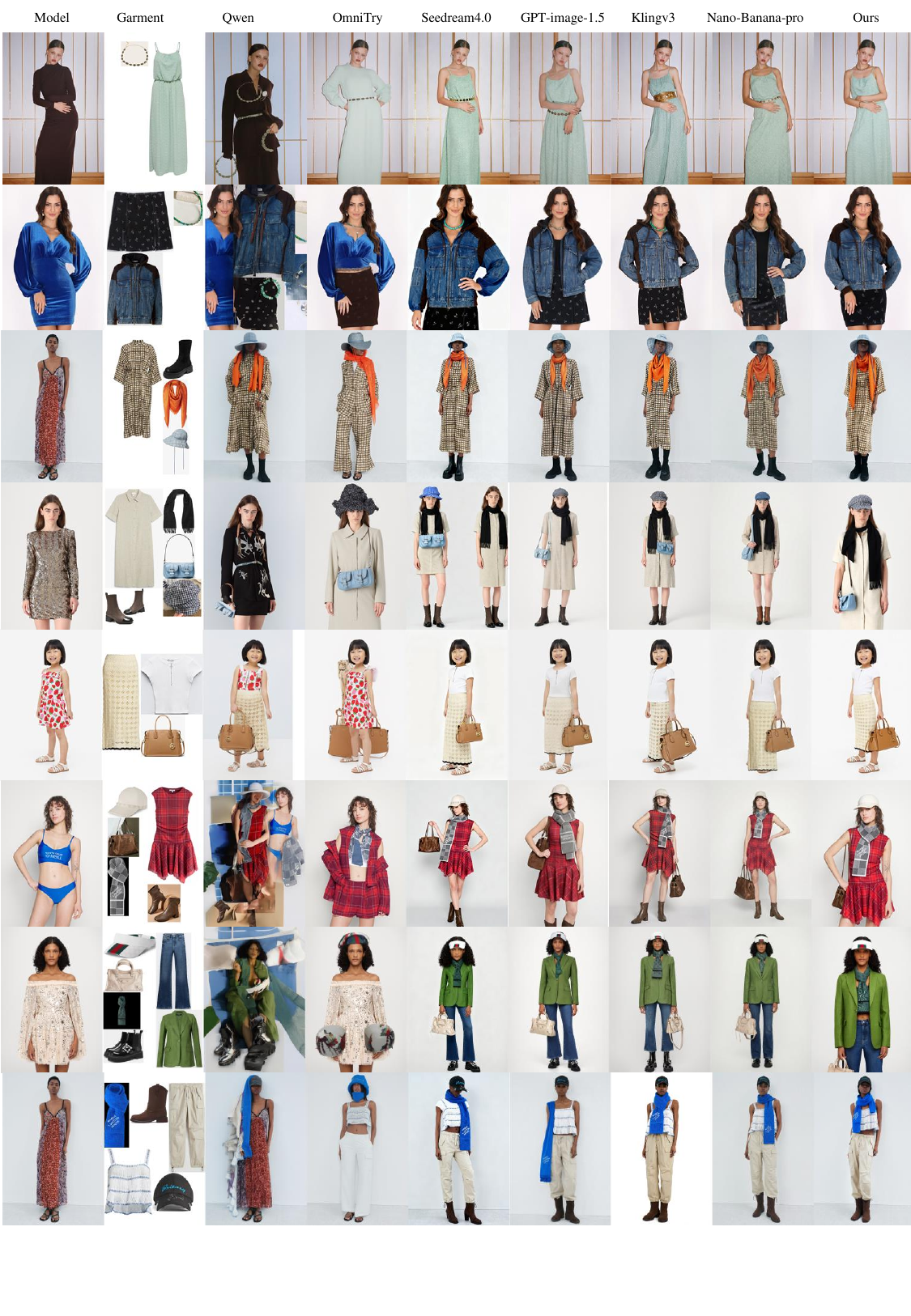}
  \caption{Qualitative comparisons on WearWow-2k.}
  \label{fig:supp1}
\end{figure}

Fig.~\ref{fig:supp1} and ~\ref{fig:supp2} display additional results of WearWow.

